\documentclass[letterpaper]{article}
\usepackage{aaai,times,helvet,courier}
\usepackage[protrusion=true,spacing=true]{microtype}
\usepackage[group-separator={,}]{siunitx}
\usepackage{amsfonts,amsthm}
\usepackage{graphicx,subfig}
\usepackage{anyfontsize}
\usepackage[absolute,overlay]{textpos}
\newtheoremstyle{mystyle}
  {5pt}		
  {0pt}		
  {}			
  {}			
  {\bfseries}	
  {.}			
  { }			
  {}			
\newtheorem{obs}{Observation}

\newcommand{\pr}{\text{Pr}}
\newcommand{\y}{\fontsize{11}{13}\selectfont}
\newcommand{\x}{\fontsize{10}{12}\selectfont\textnormal}
\newcommand{\tighteq}{\hspace{-2pt}=\hspace{-2pt}}

\newcommand{\cit}[1]{\citeauthor{#1} \citeyear{#1}}
\newcommand{\citp}[1]{\citeauthor{#1} \shortcite{#1}}

\newcommand{\plotheight}{0.181\textheight}
\newcommand{\plothpad}{20pt}
\newcommand{\plotlegheight}{0.1\textheight}

\newcommand{\matrixheight}{0.22\textheight}
\newcommand{\matrixhpad}{13pt}
\newcommand{\matrixlegheight}{0.166\textheight}

\begin{document}

	\title{An Empirical Study on the Practical Impact of Prior Beliefs over Policy Types \\[2pt]}

	\author{
		\begin{tabular}{@{}ccc@{}}
			\y{Stefano V. Albrecht} & \y{Jacob W. Crandall} & \y{Subramanian Ramamoorthy}\hspace{-1pt} \\[-2pt]
			\x{The University of Edinburgh} & \x{Masdar Institute of Science and Technology} & \x{The University of Edinburgh} \\[-2pt]
			\x{Edinburgh, United Kingdom} & \x{Abu Dhabi, United Arab Emirates} & \x{Edinburgh, United Kingdom} \\[-2.5pt]
			\x{s.v.albrecht@sms.ed.ac.uk} & \x{jcrandall@masdar.ac.ae} & \x{s.ramamoorthy@ed.ac.uk}
		\end{tabular}
	}

	\maketitle


	\begin{abstract}
		\begin{quote}
			Many multiagent applications require an agent to learn quickly how to interact with previously unknown other agents. To address this problem, researchers have studied learning algorithms which compute posterior beliefs over a hypothesised set of policies, based on the observed actions of the other agents. The posterior belief is complemented by the prior belief, which specifies the subjective likelihood of policies before any actions are observed. In this paper, we present the first comprehensive empirical study on the practical impact of prior beliefs over policies in repeated interactions. We show that prior beliefs can have a significant impact on the long-term performance of such methods, and that the magnitude of the impact depends on the depth of the planning horizon. Moreover, our results demonstrate that automatic methods can be used to compute prior beliefs with consistent performance effects. This indicates that prior beliefs could be eliminated as a manual parameter and instead be computed automatically.
		\end{quote}
	\end{abstract}

	\section{Introduction} \label{sec:intro}

Much research in AI has focused on innovative applications such as adaptive user interfaces, robotic elderly assistance, and autonomous trading agents. A key technological challenge in these applications is that an agent (virtual or physical) has to be able to learn quickly how to interact effectively with previously unknown other agents. For example, an adaptive user interface should be able to adapt effectively within minutes rather than hours. However, users may change frequently and we may not necessarily know how they behave.

Unfortunately, current methods for multiagent learning do not solve this challenge because they require long learning periods, often in the order of thousands of trials even for the simplest tasks (e.g. \cit{cb1998}). A common approach to accelerate the learning process is to introduce assumptions about the behaviour of other agents, but this may severely limit the applicability of the method. To address this limitation, researchers have studied learning systems which can utilise \emph{multiple} sets of assumptions rather than relying on one set of assumptions. The basic idea is to filter out bad assumptions based on the agents' observed behaviours.

To illustrate this idea, imagine you are playing a game of Rock-Paper-Scissors against a stranger. Our experience tells us that there is a specific set of policies which the other player could be using, such as ``play previous action of opponent'', ``play action that would have won in last round'', etc. The idea, then, is to compare the predictions of these policies with the observed actions of the player, and to plan our own actions with respect to those policies which most closely match the player's actions. There is a growing body of work showing that this method can quickly learn effective behaviours while maintaining great flexibility as to the kinds of agents it can coordinate with \cite{ar2013,bsk2011,bm2005}.

A central component of this method is the \emph{posterior belief}, which often takes the form of a probability distribution describing the relative likelihood that the agents use any of the considered policies, based on their past actions. Consequently, much research focused on the computation of posterior beliefs and convergence guarantees \cite{ar2014,gd2005}.

The posterior belief is complemented by the \emph{prior belief}, which describes the subjective likelihood of policies \emph{before} any actions are observed. However, in contrast to posterior beliefs, there are currently no systematic studies on the impact of prior beliefs in the AI literature, and it is common to use uniform priors in experiments. A reason for this may be the fact that beliefs can change rapidly after a few observations, which could suggest that prior beliefs have negligible effect.

On the other hand, there is a substantial body of work in the game theory literature arguing the importance of prior beliefs over policies, or \emph{types}, e.g. \cite{dfl2004,kl1993,j1991}. However, these works are of a theoretical nature, in the sense that they are interested in the impact of prior beliefs on equilibrium attainment. In contrast, our interest is in the \emph{practical} impact of prior beliefs, i.e. payoff maximisation. Moreover, the game theory literature on this subject almost always assumes that all players use the same Bayesian reasoning over types, while we make no such assumption.

Thus, we are left with the following questions: Do prior beliefs have an impact on the long-term payoff maximisation of the considered method? If so, how? And, importantly, can we automatically compute prior beliefs with the goal of improving the long-term performance?

This paper sets out to answer these questions. We present a comprehensive empirical study on the practical impact of prior beliefs over policies in repeated interactions. The study compares 10 automatic methods to compute prior beliefs in a range of repeated matrix games, using three different classes of automatically generated adaptive policies. Our results show that prior beliefs can indeed have a significant impact on the long-term performance of the described method. Moreover, we show that the depth of the planning horizon (i.e. how far we look into the future) plays a central role for the magnitude of the impact, and we explain how it can both diminish and amplify the impact. Finally, and perhaps most intriguingly, we show that automatic methods can compute prior beliefs that consistently optimise specific metrics (such as our own payoffs) across a variety of scenarios. An implication of this is that prior beliefs could be eliminated as a manual parameter and instead be computed automatically.

	\section{Related Work} \label{sec:relwork}

The general idea described in the previous section has been investigated by a number of researchers. As a result, there exist several models and algorithms.

Perhaps the earliest formulation was by \citp{h1967} in the form of Bayesian games. Several works analyse the dynamics of learning in Bayesian games, e.g. \cite{dfl2004,kl1993,j1991}. The empirical study by \citp{cjm2001} is close in spirit to our work. However, their focus is on equilibrium attainment rather than payoff maximisation.

\citp{ar2013} presented an algorithm, called HBA, which utilises a set of hypothesised types (or policies) in a planning procedure to find optimal responses. Related algorithms were studied by \citp{bsk2011} and \citp{cm1999}.

\citp{gd2005} study types under partially observable system states, in a model called I-POMDP. Their type specifications include additional structure such as complex nested beliefs and subjective optimality criteria.

\citp{bm2005} describe an algorithm which uses ``play books'' to plan its actions. Each ``play'' in a play book specifies a complete team policy, with additional structure such as roles for each agent in the team.

Many algorithms for plan recognition use a similar concept in the form of ``plan libraries'' \cite{c2001,cg1993}. The idea is to learn the goal of an agent by comparing its actions with the plans in the library.

All of these methods use some form of a prior belief over action policies. Our work can be viewed as complementing these works by showing how prior beliefs can affect the long-term performance of such methods.

Finally, there is a substantial body of work on \text{``uninformed''} priors (e.g. \citp{b1979}, \citp{j1968}, and references therein). The purpose of such priors is to express a state of complete uncertainty, whilst possibly incorporating subjective prior information. (What this means and whether this is possible has been the subject of a long debate, e.g. \cit{f2008}.) This is different from our priors, which are shaped with the goal of payoff maximisation rather than expressing uncertainty or incorporating prior information.

	\section{Experimental Setup} \label{sec:expsetup}

This section describes the experimental setup used in our study. All parameter settings can be found in an appendix document \cite{acr2015priorapp}.

		\subsection{Games}

We used a comprehensive set of benchmark games introduced by \citp{rg1966}, which consists of 78 repeated $2 \times 2$ matrix games (i.e. 2 players with 2 actions). The games are \emph{strictly ordinal}, meaning that each player ranks each of the 4 possible outcomes from 1 (least preferred) to 4 (most preferred), and no two outcomes have the same rank. Furthermore, the games are \emph{distinct} in the sense that no game can be obtained by transformation of any other game, which includes interchanging the rows, columns, and players (and any combination thereof) in the payoff matrix of the game.

The games can be grouped into 21 \emph{no-conflict} games and 57 \emph{conflict} games. In a no-conflict game, the two players have the same most preferred outcome, and so it is relatively easy to arrive at a solution that is best for both players. In a conflict game, the players disagree on the best outcome, hence they will have to find some form of a compromise.

		\subsection{Performance Criteria}

Each play of a game was partitioned into \emph{time slices} which consist of an equal number of consecutive time steps. For each time slice, we measured the following performance criteria:

\begin{description}
	\item[Convergence:] An agent converged in a time slice if its action probabilities in the time slice did not deviate by more than 0.05 from its initial action probabilities in the time slice. Returns 1 (true) or 0 (false) for each agent.
	\item[Average payoff:] Average of payoffs an agent received in the time slice. Returns value in $[1,4]$ for each agent.
	\item[Welfare and fairness:] Average sum and product, respectively, of the joint payoffs received in the time slice. Returns values in $[2,8]$ and $[1,16]$, respectively.
	\item[Game solutions:] Tests if the averaged action probabilities of the agents formed an approximate stage-game Nash equilibrium, Pareto optimum, Welfare optimum, or Fairness optimum in the time slice. Returns 1 (true) or 0 (false) for each game solution.
\end{description}

Precise formal definitions of these performance criteria can be found in \cite{ar2012}.

		\subsection{Algorithm} \label{sec:hba}

We used \emph{Harsanyi-Bellman Ad Hoc Coordination} (HBA) \cite{ar2014} as a canonical example of the idea outlined in Section~\ref{sec:intro}. In the following, HBA controls player $i$ while $j$ denotes the other player. HBA uses a set $\Theta_j^*$ of \emph{types} which can be provided by a human expert or, as we do in this work, generated automatically (see Section~\ref{sec:types}). We assume that the true type of player $j$ is always included in $\Theta_j^*$, but we do not know which type in $\Theta_j^*$ it is.

Each type $\vspace{-1pt}\theta_j^* \hspace{-1pt}\in\hspace{-1pt} \Theta_j^*$ specifies a complete action policy for player $j$. We write $\pi_j(H^t,a_j,\theta_j^*)$ to denote the probability that player $j$ chooses action $a_j$ if it is of type $\theta_j^*$, given the history $H^t$ of previous joint actions. HBA computes the posterior belief that player $j$ is of type $\theta_j^*$, given $H^t$, as
\begin{displaymath}
	\pr_j(\theta_j^* | H^t) = \eta \, P_j(\theta_j^*) \prod_{\tau=0}^{t-1} \pi_j(H^{\tau}\hspace{-1pt},a_j^\tau,\theta_j^*)
\end{displaymath}
where $\eta$ is a normalisation constant and $P_j(\theta_j^*)$ denotes the \emph{prior belief} that player $j$ is of type $\theta_j^*$.

Using the posterior $\pr_j$, HBA chooses an action $a_i$ which maximises the expected payoff $E_h^{a_i}(H^t)$, defined as
\begin{displaymath}
	E_h^{a_i}(\hat{H}) = \hspace{-4pt} \sum_{\theta_j^* \in \Theta_j^*} \hspace{-3pt} \pr_j(\theta_j^* | H^t) \hspace{-2pt} \sum_{a_j \in A_j} \hspace{-3pt} \pi_j(\hat{H},a_j,\theta_j^*) \, Q_{h-1}^{(a_i,a_j)}(\hat{H})
\end{displaymath}
\begin{displaymath}
	Q_h^{(a_i,a_j)}(\hat{H}) = u_i(a_i,a_j) + \left\{ \begin{array}{l} 0 \ \ \text{if}\ \ h = 0, \ \text{else} \\ \max_{a'_i} E_h^{a'_i} \hspace{-2pt} \left( \langle \hat{H},(a_i,a_j)\rangle \right) \end{array} \right.
\end{displaymath}
where $u_i(a_i,a_j)$ is player $i$'s payoff if joint action $\vspace{1pt}(a_i,a_j)$ is played, and $h$ specifies the depth of the planning horizon (i.e. HBA predicts the next $h$ actions of player $j$). Note that $H^t$ is the current history of joint actions while $\hat{H}$ is used to construct all future trajectories in the game.

		\subsection{Types} \label{sec:types}

We used three methods to generate parameterised sets of types $\Theta_j^*$ for a given game. The generated types cover a reasonable spectrum of adaptive behaviours, including deterministic (CDT), randomised (CNN), and hybrid (LFT) policies.

			\subsubsection{Leader-Follower-Trigger Agents (LFT)} \label{sec:lfg}

\citp{c2014} described a method to automatically generate sets of ``leader'' and ``follower'' agents that seek to play specific sequences of joint actions, called ``target solutions''. A leader agent plays its part of the target solution as long as the other player does. If the other player deviates, the leader agent punishes the player by playing a minimax strategy. The follower agent is similar except that it does not punish. Rather, if the other player deviates, the follower agent randomly resets its position within the target solution and continues play as usual. We augmented this set by a trigger agent which is similar to the leader and follower agents, except that it plays its maximin strategy indefinitely once the other player deviates.

			\subsubsection{Co-Evolved Decision Trees (CDT)}

We used genetic programming \cite{k1992} to automatically breed sets of decision trees. A decision tree takes as input the past $n$ actions of the other player (in our case, $n = 3$) and deterministically returns an action to be played in response. The breeding process is co-evolutional, meaning that two pools of trees are bred concurrently (one for each player). In each evolution, a random selection of the trees for player 1 is evaluated against a random selection of the trees for player 2. The fitness criterion includes the payoffs generated by a tree as well as its dissimilarity to other trees in the same pool. This was done to encourage a more diverse breeding of trees, as otherwise the trees tend to become very similar or identical.

			\subsubsection{Co-Evolved Neural Networks (CNN)}

We used a string-based genetic algorithm \cite{h1975} to breed sets of artificial neural networks. The process is basically the same as the one used for decision trees. However, the difference is that artificial neural networks can learn to play stochastic strategies while decision trees always play deterministic strategies. Our networks consist of one input layer with 4 nodes (one for each of the two previous actions of both players), a hidden layer with 5 nodes, and an output layer with 1 node. The node in the output layer specifies the probability of choosing action 1 (and, since we play $2 \times 2$ games, of action 2). All nodes use a sigmoidal threshold function and are fully connected to the nodes in the next layer.

		\subsection{Prior Beliefs} \label{sec:priors}

We specified a total of 10 different methods to automatically compute prior beliefs $P_j$ for a given set of types $\Theta_j^*$:

			\subsubsection{Uniform prior}

The uniform prior sets $P_j(\theta_j^*) = |\Theta_j^*|^{-1}$ for all $\theta_j^* \in \Theta_j^*$. This is the baseline prior against which the other priors are compared.

			\subsubsection{Random prior}

The random prior specifies $P_j(\theta_j^*) = .0001$ for half of the types in $\Theta_j^*$ (selected at random). The remaining probability mass is uniformly spread over the other half. The random prior is used to check if the performance differences of the various priors may be purely due to the fact that they concentrate the probability mass on fewer types.

			\subsubsection{Value priors}

Let $U_k^t(\theta_j^*)$ be the expected cumulative payoff to player $k$, from the start up until time $t$, if player $j$ (i.e. the other player) is of type $\theta_j^*$ and player $i$ (i.e. HBA) plays optimally against it. Each value prior is in the general form of $P_j(\theta_j^*) = \eta \, \psi(\theta_j^*)^b$, where $\eta$ is a normalisation constant and $b$ is a ``booster'' exponent used to magnify the differences between types $\theta_j^*$ (in this work, we use $b = 10$). Based on this general form, we define four different value priors:


\begin{itemize}
	\item Utility prior: $\psi_U(\theta_j^*) = U^t_i(\theta_j^*)$
	\item Stackelberg prior: $\psi_S(\theta_j^*) = U^t_j(\theta_j^*)$
	\item Welfare prior: $\psi_W(\theta_j^*) = U^t_i(\theta_j^*) + U^t_j(\theta_j^*)$
	\item Fairness prior: $\psi_F(\theta_j^*) = U^t_i(\theta_j^*) * U^t_j(\theta_j^*)$
\end{itemize}

Our choice of value priors is motivated by the variety of metrics they cover. As a result, these priors can produce substantially different probabilities for the same set of types.

\begin{figure*}[ht]
	\vspace{-10pt}
	\centering
	\subfloat[LFT\,($h \tighteq 1$) -- CFP -- No-Conflict]{\includegraphics[height=\plotheight]{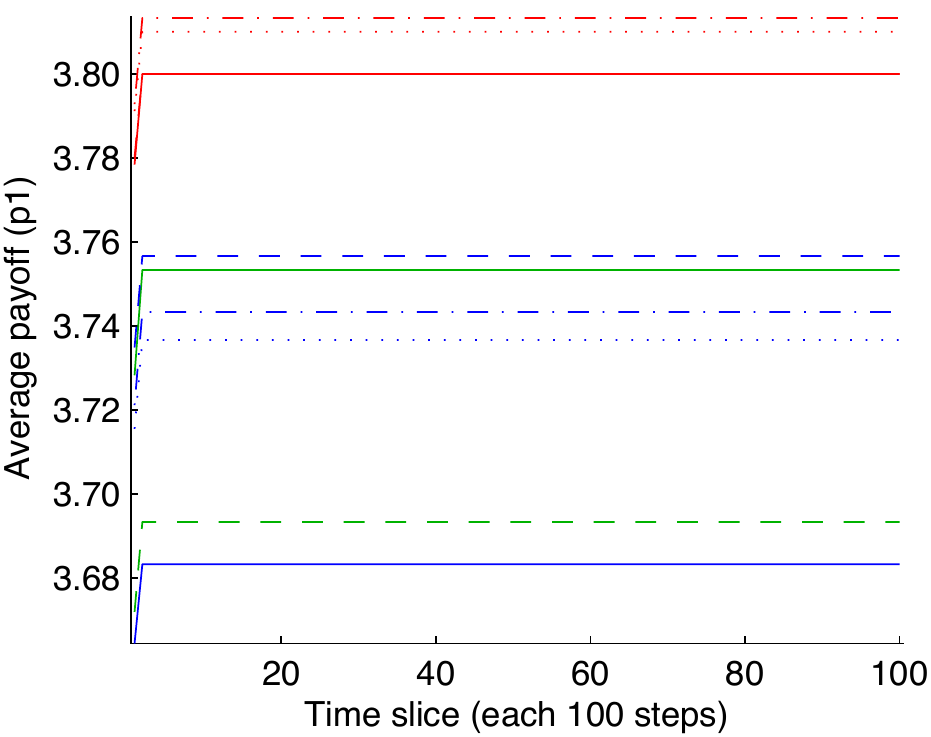}}\hspace{\plothpad}
	\subfloat[CDT\,($h \tighteq 3$) -- FP -- Conflict]{\includegraphics[height=\plotheight]{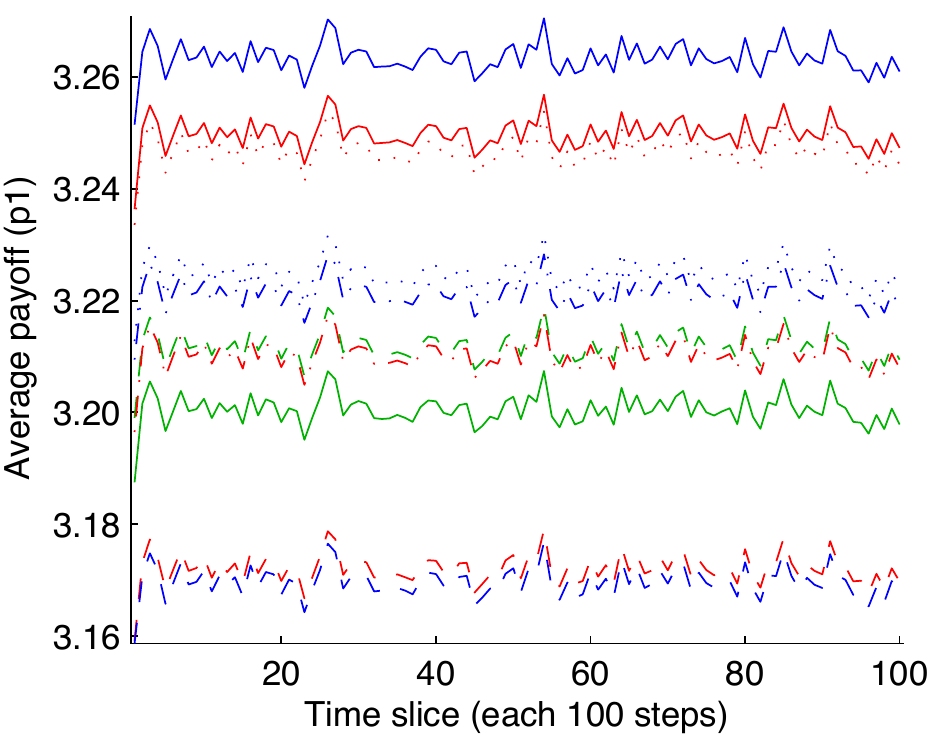}}\hspace{\plothpad}
	\subfloat[CNN\,($h \tighteq 5$) -- FP -- Conflict]{\includegraphics[height=\plotheight]{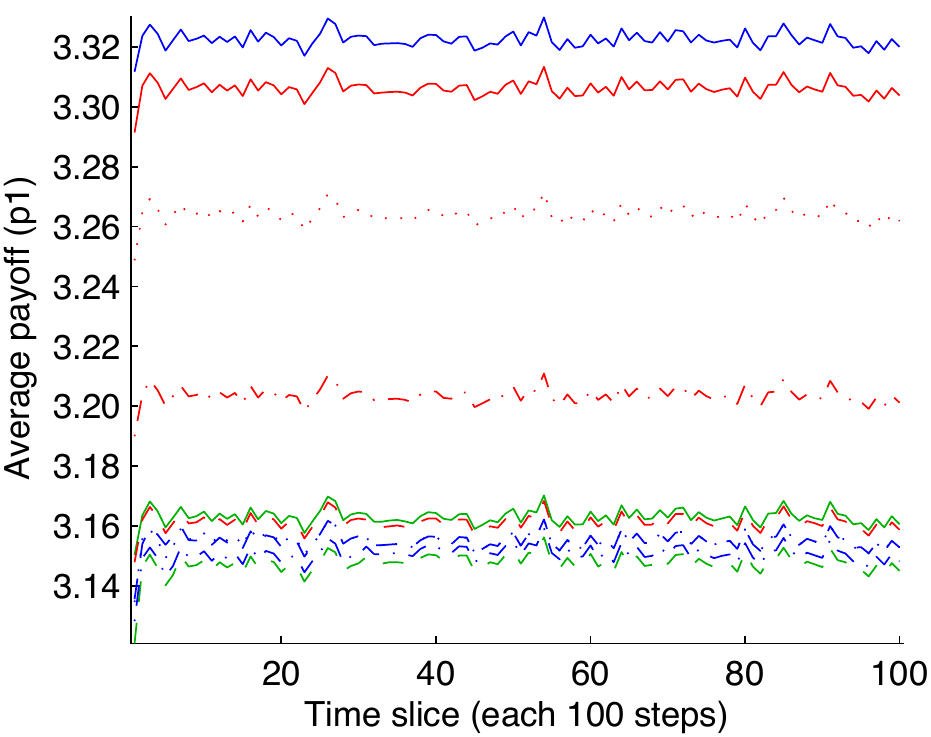}}
	\caption{\textbf{Prior beliefs can have significant impact on long-term performance.} Plots show average payoffs of player 1 (HBA). X($h$)--Y--Z format: HBA used X types and horizon $h$, player 2 was controlled by Y, and results are averaged over Z games.}
	\label{fig:impact}
\end{figure*}

			\subsubsection{LP-priors} \label{sec:lp-priors}

LP-priors are based on the idea that optimal priors can be formulated as the solution to a mathematical optimisation problem (in this case, a linear programme). Each LP-prior generates a quadratic matrix $A$, where each element $A_{j,j'}$ contains the ``loss'' that HBA would incur if it planned its actions against the type $\theta_{j'}^*$ while the true type of player $j$ is $\theta_j^*$. Formally, let $U_k^t(\theta_j^*|\theta_{j'}^*)$ be like $U_k^t(\theta_j^*)$ except that HBA thinks that player $j$ is of type $\theta_{j'}^*$ instead of $\theta_j^*$. We define four different LP-priors:

\begin{itemize}
	\item LP-Utility: $A_{j,j'} = \psi_U(\theta_j^*) - U^t_i(\theta_j^* | \theta_{j'}^*)$
	\item LP-Stackelberg: $A_{j,j'} = \psi_S(\theta_j^*) - U^t_j(\theta_j^* | \theta_{j'}^*)$
	\item LP-Welfare: $A_{j,j'} = \psi_W(\theta_j^*) - [U^t_i(\theta_j^* | \theta_{j'}^*) + U^t_j(\theta_j^* | \theta_{j'}^*)]$
	\item LP-Fairness: $A_{j,j'} = \psi_F(\theta_j^*) - [U^t_i(\theta_j^* | \theta_{j'}^*) * U^t_j(\theta_j^* | \theta_{j'}^*)]$
\end{itemize}
 
The matrix $A$ can be fed into a linear program of the form $\min_c c^Tx$ s.t. $[z,A] x \leq 0$, with $n = |\Theta_j^*|$, $c = (1,\left\{ 0 \right\}^n)^T$, $z = (\left\{ -1 \right\}^n)^T$, to find a vector $x = (l,p_1,...,p_n)$ in which $l$ is the minimised expected loss to HBA when using the probabilities $p_1,...,p_n$ (one for each type) as the prior belief $P_j$. In order to avoid premature elimination of types, we furthermore require that $p_v > 0$ for all $1 \leq v \leq n$.

While this is a mathematically rigorous formulation, it is important to note that it is a simplification of how HBA really works. HBA uses the prior in every recursion of its planning procedure, while the LP formulation implicitly assumes that HBA uses the prior to randomly sample one of the types against which it then plans optimally. Nonetheless, this is often a reasonable approximation.

		\subsection{Experimental Procedure}

We performed identical experiments for every type generation method described in Section~\ref{sec:types}. Each of the 78 games was played 10 times with different random seeds, and each play was repeated against three opponents (30 plays in total):

\begin{description}
	\item[1. (RT)] A randomly generated type was used to control player 2 and the play lasted 100 rounds.
	\item[2. (FP)] A fictitious player \cite{b1951} was used to control player 2 and the play lasted 10,000 rounds.
	\item[3. (CFP)] A conditioned fictitious player (which learns action distributions conditioned on the previous joint action) was used to control player 2 and the play lasted 10,000 rounds.
\end{description}

In each play, we randomly generated 9 unique types and provided them to HBA along with the true type of player 2, such that $|\Theta_2^*| = 10$. Thus, in each play there was a time after which HBA knew the type of player 2. To avoid ``end-game'' effects, the players were unaware the number of rounds.

We included FP and CFP because they try to learn the behaviour of HBA. (While the generated types are adaptive, they do not create models of HBA's behaviour.) To facilitate the learning, we allowed for 10,000 rounds. Finally, since FP and CFP will always choose dominating actions if they exist (in which case there is no interaction), we filtered out all games in the FP and CFP plays that had a dominating action for player 2 (leaving 15 no-conflict and 33 conflict games).

\begin{textblock}{20}(48.8,9)
	\includegraphics[height=\plotlegheight]{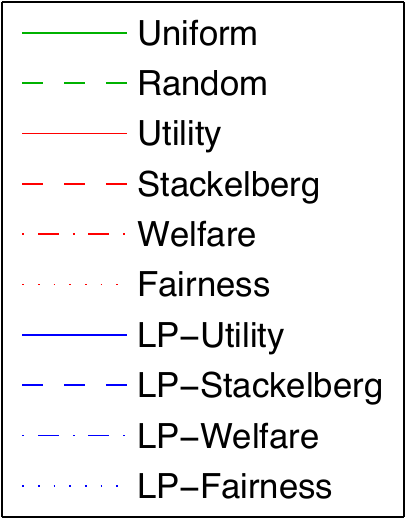}
\end{textblock}

\begin{figure*}[ht]
	\vspace{-10pt}
	\centering
	\subfloat[$h = 1$]{\includegraphics[height=\plotheight]{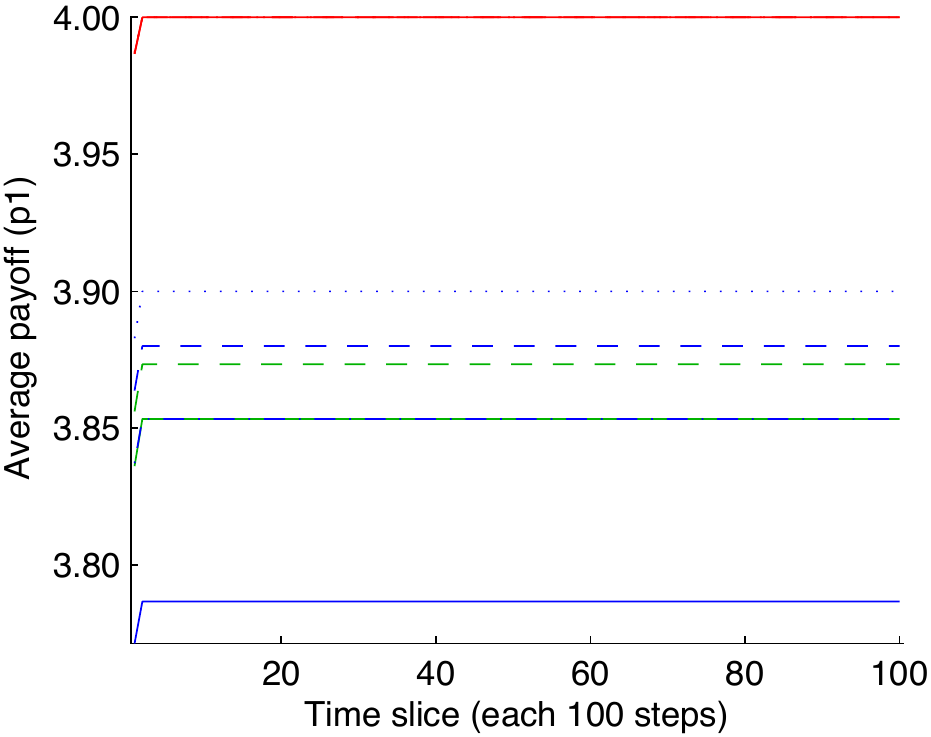}}\hspace{\plothpad}
	\subfloat[$h = 3$]{\includegraphics[height=\plotheight]{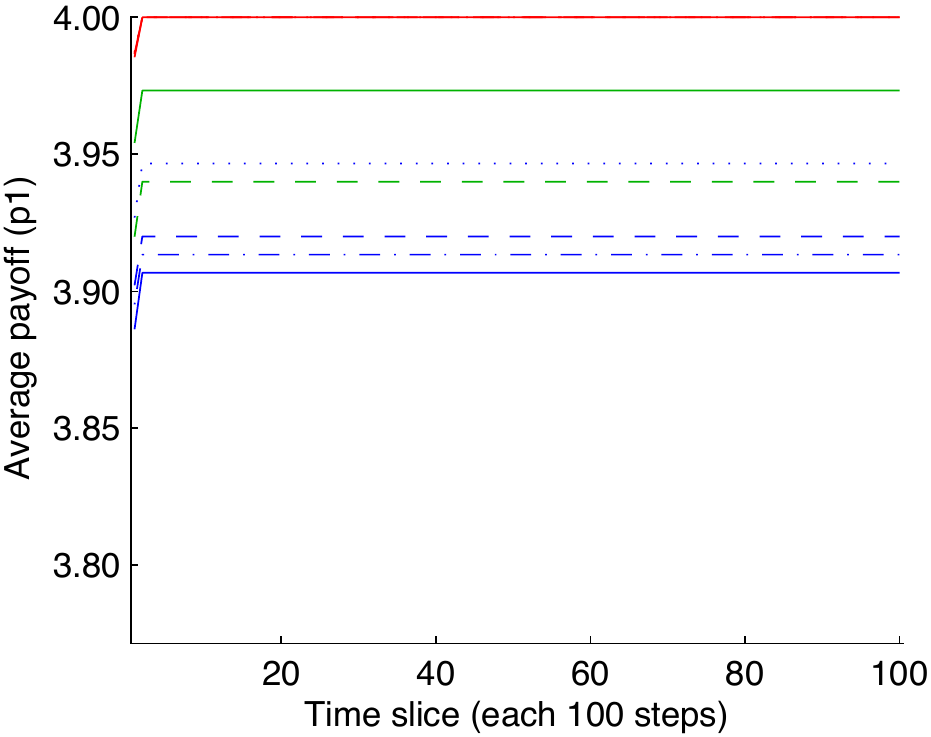}}\hspace{\plothpad}
	\subfloat[$h = 5$]{\includegraphics[height=\plotheight]{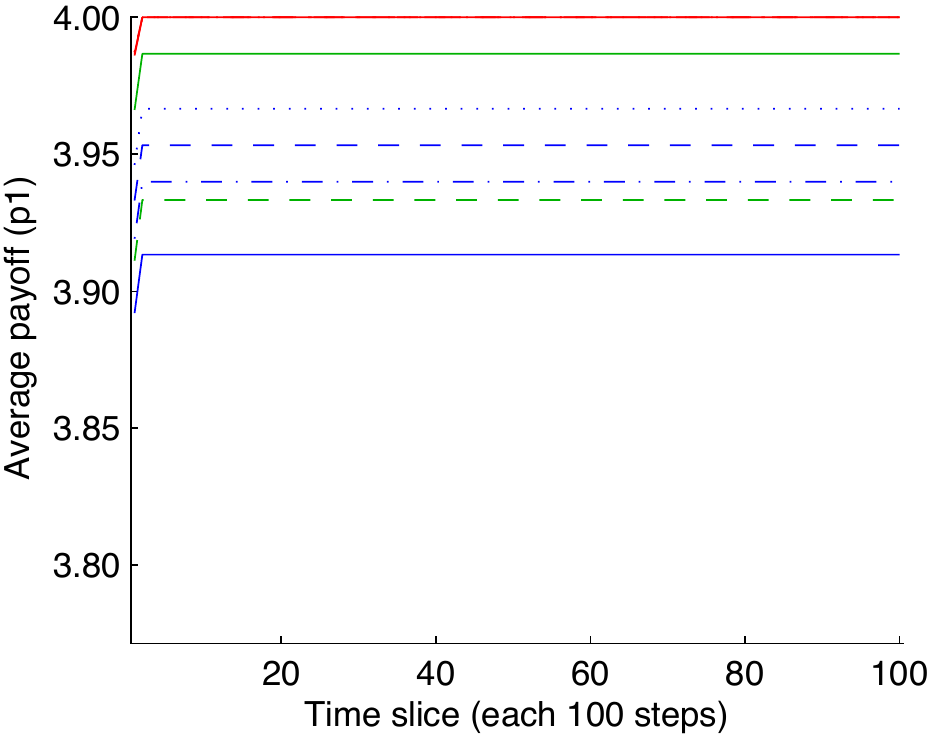}}
	\caption{\textbf{Deeper planning horizons can diminish impact of prior beliefs.} Results shown for HBA with LFT types, player 2 controlled by FP, averaged over no-conflict games. $h$ is depth of planning horizon (i.e. predicting $h$ next actions of player 2).}
	\label{fig:diminish}
\end{figure*}

\begin{figure*}[ht]
	\vspace{-10pt}
	\centering
	\subfloat[$h = 1$]{\includegraphics[height=\plotheight]{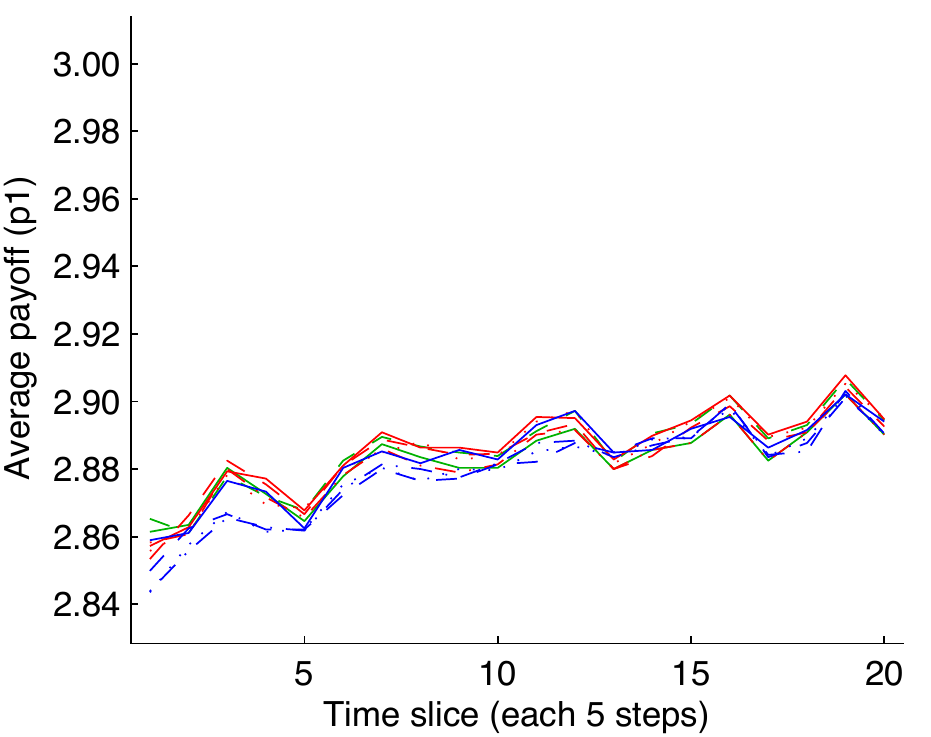}}\hspace{\plothpad}
	\subfloat[$h = 3$]{\includegraphics[height=\plotheight]{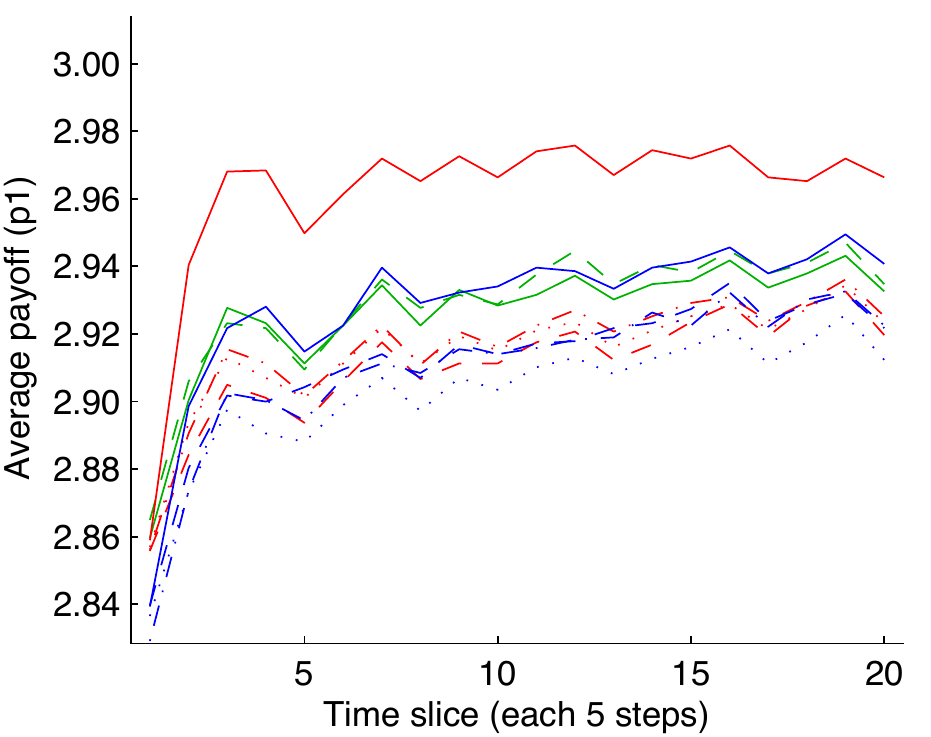}}\hspace{\plothpad}
	\subfloat[$h = 5$]{\includegraphics[height=\plotheight]{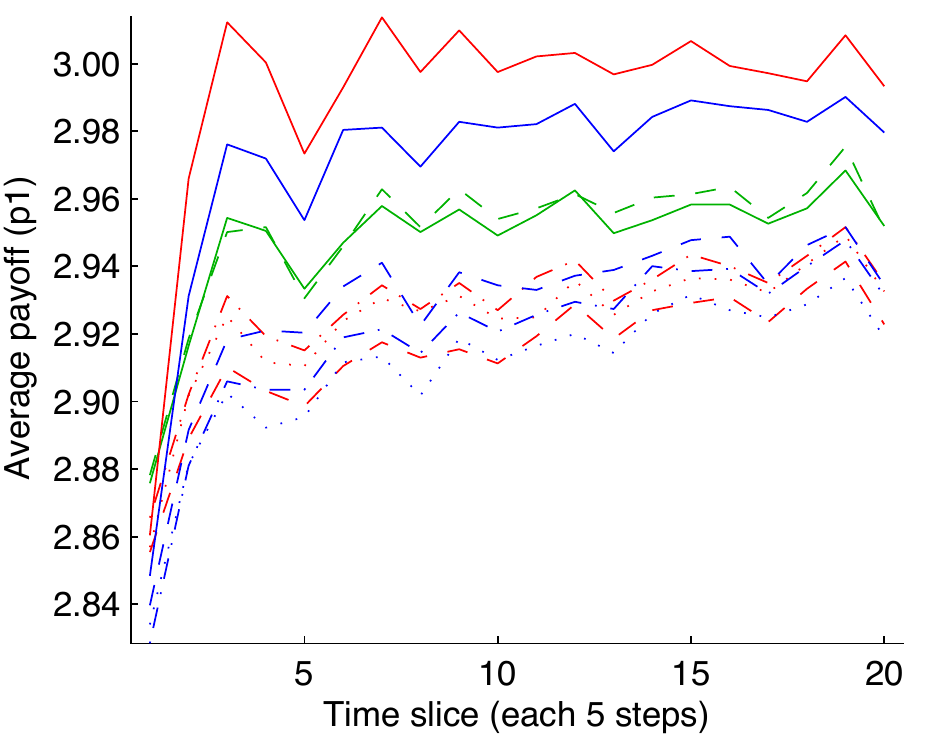}}
	\caption{\textbf{Deeper planning horizons can amplify impact of prior beliefs.} Results shown for HBA with CNN types, player 2 controlled by RT, averaged over conflict games. $h$ is depth of planning horizon (i.e. predicting $h$ next actions of player 2).}
	\label{fig:amplify}
\end{figure*}

	\section{Results} \label{sec:results}

All experimental results can be found in the appendix document \cite{acr2015priorapp}. Based on this data, we make three main observations:

\begin{obs} \label{obs:1}
	Prior beliefs can have a significant impact on the long-term performance of HBA.
\end{obs}

This was observed in all classes of types, against all classes of opponents, and in all classes of games used in this study. Figure~\ref{fig:impact} provides three representative examples from a range of scenarios. Many of the relative differences due to prior beliefs were statistically significant, based on paired two-sided t-tests with a 5\% significance level.

Our data explain this as follows: Different prior beliefs may cause HBA to take different actions at the beginning of the game. These actions will shape the beliefs of the other player (i.e. how it models and adapts to HBA's actions) which in turn will affect HBA's next actions. Thus, if different prior beliefs lead to different initial actions, they may lead to different play trajectories with different payoffs.

Given that there is a time after which HBA will know the true type of player 2 (since it is provided to HBA), it may seem surprising that this process would lead to differences in the long-term. In fact, in our experiments, HBA often learned the true type after only 3 to 5 rounds, and in most cases in under 20 rounds. After that point, if the planning horizon of HBA is sufficiently deep, it will realise if its initial actions were sub-optimal and if it can manipulate the play trajectory to achieve higher payoffs in the long-term, thus diminishing the impact of prior beliefs.

However, deep planning horizons can be problematic in practice since the time complexity of HBA is exponential in the depth of the planning horizon. Therefore, the planning horizon constitutes a trade-off between decision quality and computational tractability. Interestingly, our data show that if we increase the depth, but stay below a sufficient depth (``sufficient'' as described above), it may also \emph{amplify} the impact of prior beliefs:

\begin{obs} \label{obs:2}
	Deeper planning horizons can both diminish and amplify the impact of prior beliefs.
\end{obs}

Again, this was observed in all tested scenarios. Figures~\ref{fig:diminish} and \ref{fig:amplify} show examples in which deeper planning horizons diminish and amplify the impact of prior beliefs, respectively.

How can deeper planning horizons amplify the impact of prior beliefs? Our data show that whether or not different prior beliefs cause HBA to take different initial actions depends not only on the prior beliefs and types, but also on the depth of the planning horizon. In some cases, differences between types (i.e. in their action choices) may be less visible in the near future and more visible in the distant future. In such cases, an HBA agent with a myopic planning horizon may choose the same (or similar) initial actions, despite different prior beliefs, because the differences in the types may not be visible within its planning horizon. On the other hand, an HBA agent with a deeper planning horizon may see the differences between the types and decide to choose different initial actions based on the prior beliefs.

We now turn to a comparison between the different prior beliefs. Here, our data reveal an intriguing property:

\begin{obs} \label{obs:3}
	Automatic methods can compute prior beliefs with consistent performance effects.
\end{obs}

\begin{figure*}[ht]
	\vspace{-10pt}
	\centering
	\subfloat[LFT\,($h \tighteq 5$) -- RT -- Conflict]{\includegraphics[height=\matrixheight]{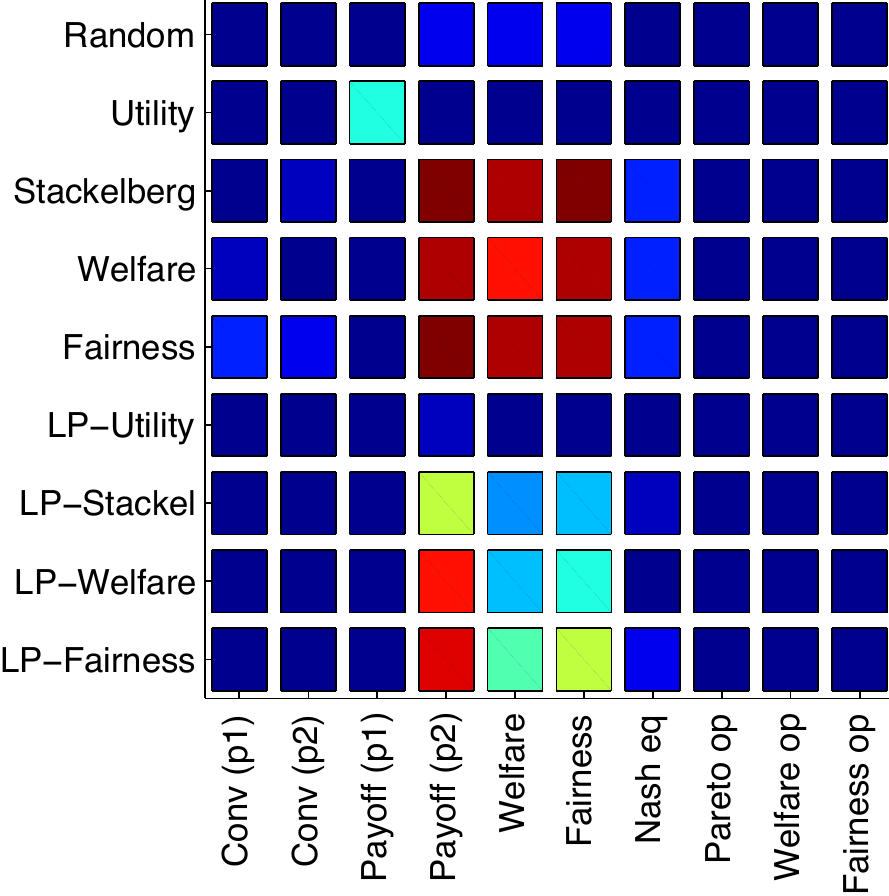}}\hspace{\matrixhpad}
	\subfloat[LFT\,($h \tighteq 5$) -- CFP -- Conflict]{\includegraphics[height=\matrixheight]{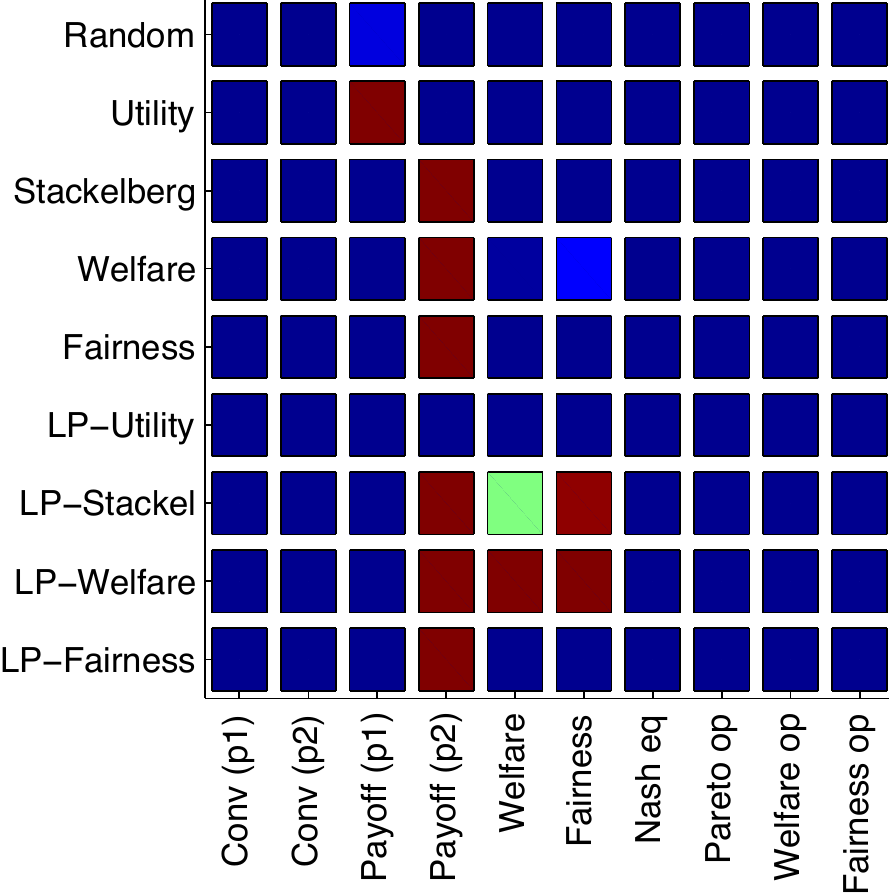}}\hspace{\matrixhpad}
	\subfloat[CDT\,($h \tighteq 3$) -- RT -- Conflict]{\includegraphics[height=\matrixheight]{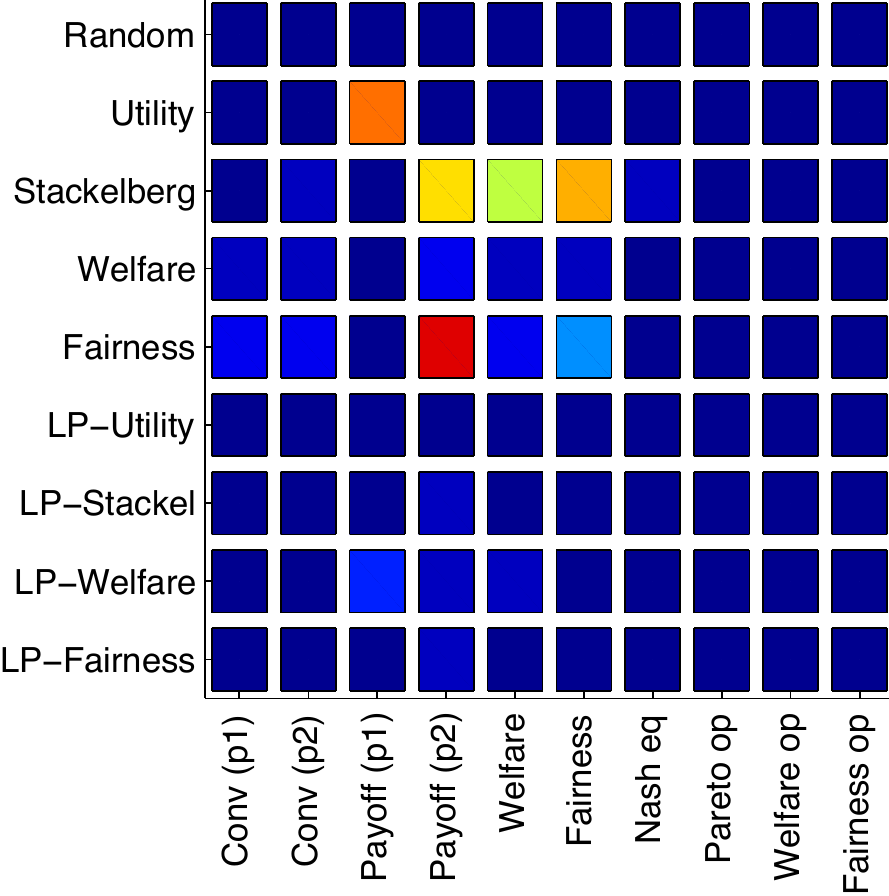}}\hspace{\matrixhpad}
	\subfloat{\raisebox{33pt}{\includegraphics[height=\matrixlegheight]{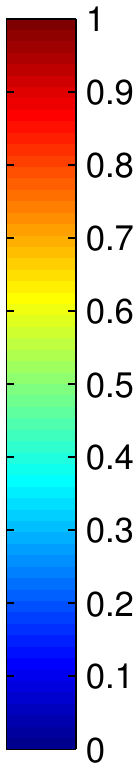}}}\\
	\subfloat[CDT\,($h \tighteq 3$) -- FP -- Conflict]{\includegraphics[height=\matrixheight]{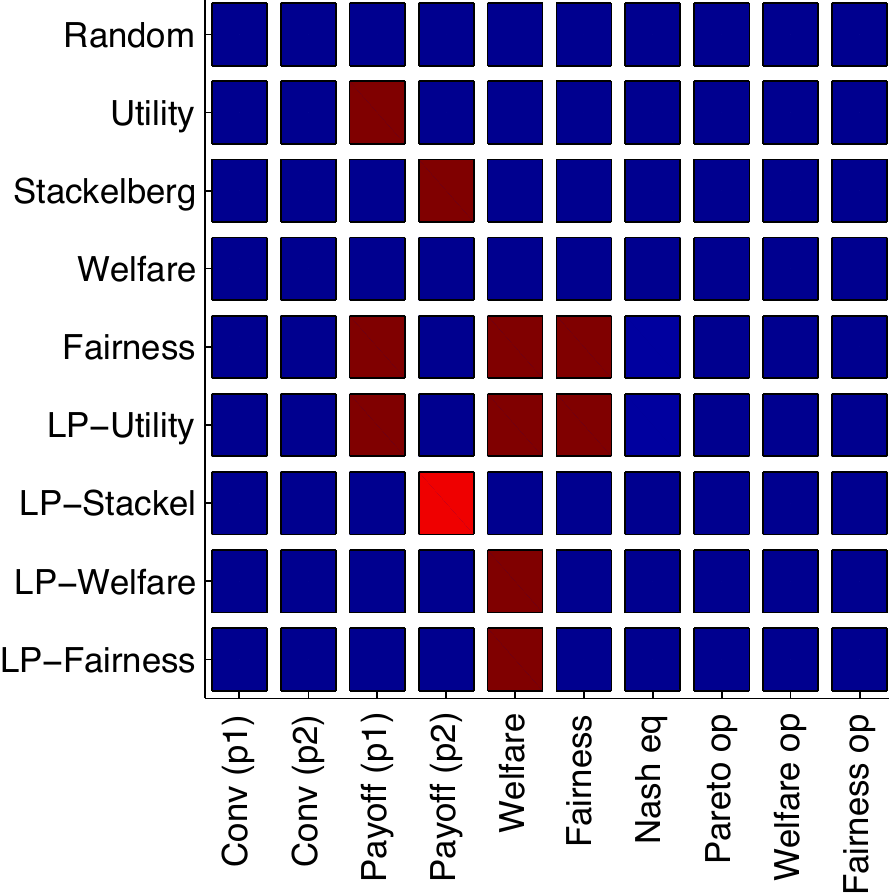}}\hspace{\matrixhpad}
	\subfloat[CNN\,($h \tighteq 3$) -- RT -- Conflict]{\includegraphics[height=\matrixheight]{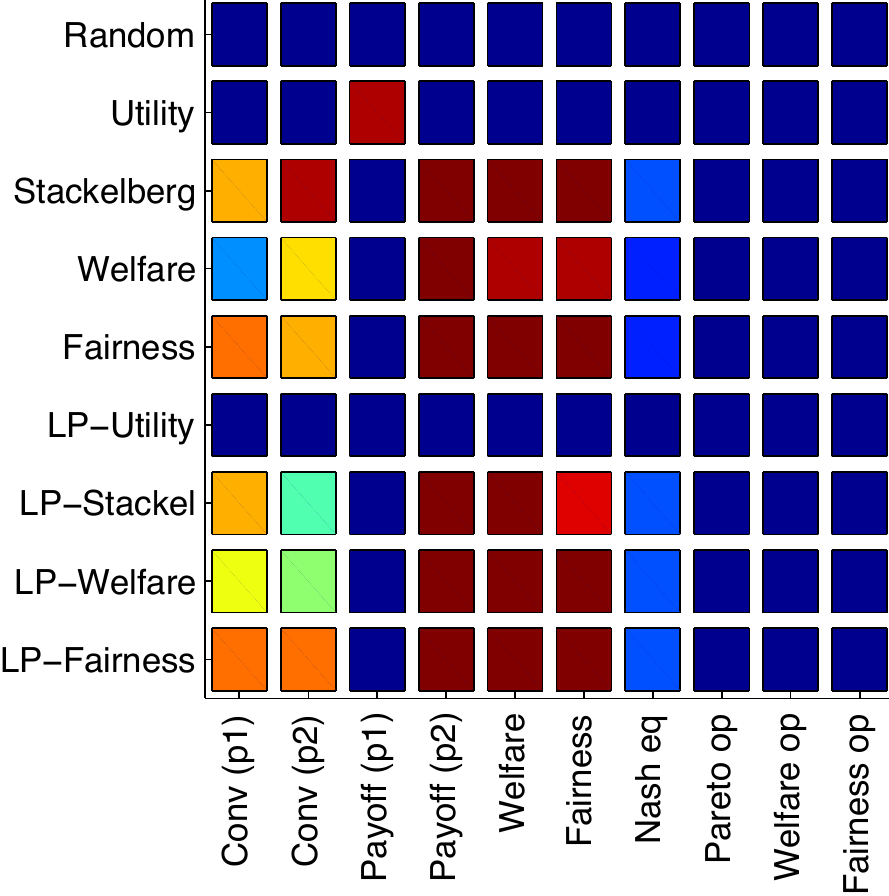}}\hspace{\matrixhpad}
	\subfloat[CNN\,($h \tighteq 1$) -- FP -- Conflict]{\includegraphics[height=\matrixheight]{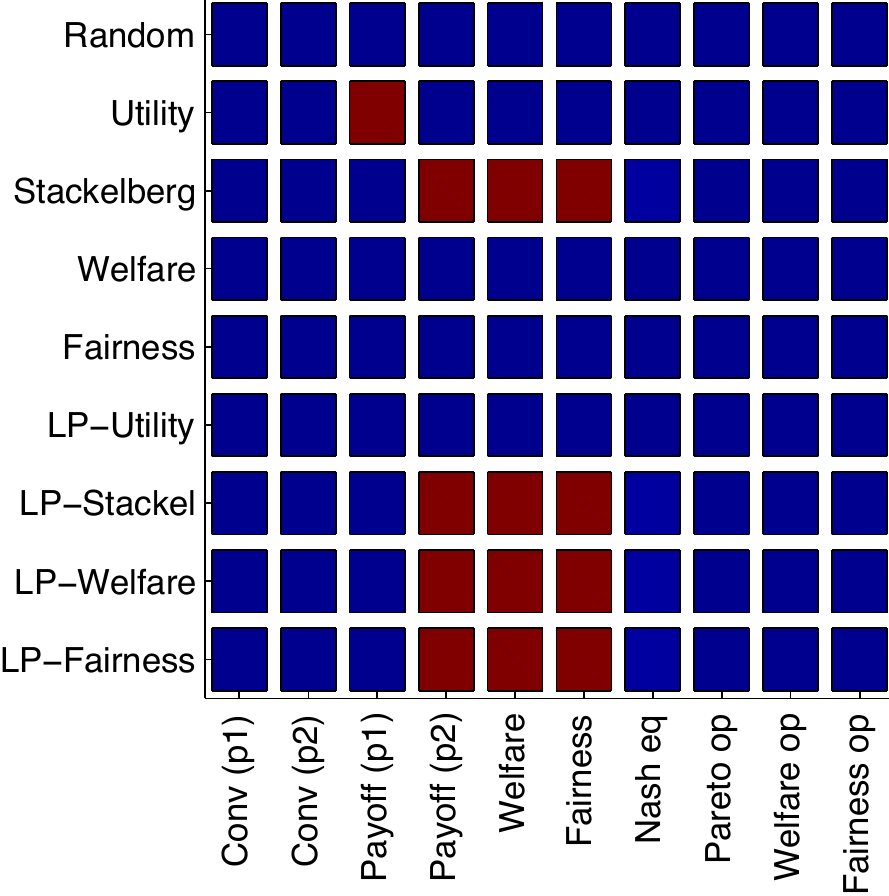}}\hspace{\matrixhpad}
	\subfloat{\raisebox{33pt}{\includegraphics[height=\matrixlegheight]{colorbar.pdf}}}
	\caption{\textbf{Automatic prior beliefs have consistent performance effects.} Rows show prior beliefs and columns show performance criteria. Each element $(r,c)$ in the matrix corresponds to the percentage of time slices in which the prior belief $r$ produced significantly higher values for the criterion $c$ than the Uniform prior, averaged over all plays in all tested games. All significance statements are based on paired right-sided t-tests with a 5\% significance level. See Figure~\ref{fig:impact} for X($h$)--Y--Z format.}
	\label{fig:consistency}
\end{figure*}

Figure~\ref{fig:consistency} shows that the prior beliefs had consistent performance effects across a wide variety of scenarios. For example, the Utility prior produced consistently higher payoffs for player 1 (i.e. HBA) while the Stackelberg prior produced consistently higher payoffs for player 2 as well as higher welfare and fairness. The Welfare and Fairness priors were similar to the Stackelberg prior, but not quite as consistent. Similar results were observed for the LP variants of the priors, despite the fact that the LP formulation is a simplification of how HBA works (cf. Section~\ref{sec:lp-priors}).

We note that none of the prior beliefs, including the Uniform prior, produced high rates for the game solutions (i.e. Nash equilibrium, Pareto optimality, etc.). This is because we measured \emph{stage-game} solutions, which have no notion of time. These can be hard to attain in repeated games, especially if the other player does not actively seek a specific solution, as was often the case in our study.

\begin{textblock}{20}(48.8,9)
	\includegraphics[height=\plotlegheight]{legend.pdf}
\end{textblock}

\begin{textblock}{20}(48.8,26.55)
	\includegraphics[height=\plotlegheight]{legend.pdf}
\end{textblock}

Observation~\ref{obs:3} is intriguing because it indicates that prior beliefs could be eliminated as a manual parameter and instead be computed automatically, using methods such as the ones specified in Section~\ref{sec:priors}. The fact that our methods produced consistent results means that prior beliefs can be constructed to optimise specific performance criteria. Note that this result is particularly interesting because the prior beliefs have no influence, whatsoever, on the true type of player 2.

This observation is further supported by the fact that the Random prior did not produce consistently different values (for any criterion) from the Uniform prior. This means that the differences in the prior beliefs are not merely due to the fact that they concentrate the probability mass on fewer types, but rather that the prior beliefs reflect the intrinsic metrics based on which they are computed (e.g. player 1's payoffs for Utility prior, player 2's payoffs for Stackelberg prior, etc.).

How is this phenomenon explained? We believe this may be an interesting analogy to the ``optimism in uncertainty'' principle (e.g. \cit{bt2003}). The optimism lies in the fact that HBA commits to a specific class of types -- those with high prior belief -- while, in truth and without further evidence, there is no reason to believe that any one type is more likely than others.

Each class of types is characterised by the intrinsic metric of the prior belief. For instance, the Utility prior assigns high probability to those types which would yield high payoffs to HBA if it played optimally against them. By committing to such a characterisation, HBA can effectively utilise Observation~\ref{obs:1} by choosing initial actions so as to shape the interaction to maximise the intrinsic metric. If the true type of player 2 is indeed this class of types, then the interaction will proceed as planned by HBA and the intrinsic metric will be optimised. However, if the true type is not in this class, then HBA will quickly learn the correct type and adjust its play accordingly, though without necessarily maximising the intrinsic metric.

This is in contrast to the Uniform and Random priors, which have no intrinsic metric. Under these priors, HBA will plan its actions with respect to types which are not characterised by a common theme (i.e., all types under the Uniform prior, and a random half under the Random prior). Therefore, HBA cannot effectively utilise Observation~\ref{obs:1}.

	\section{Conclusion} \label{sec:concl}

This paper presents the first comprehensive empirical study on the practical impact of prior beliefs over policy types in repeated interactions. The three messages to take away are: (1) prior beliefs can have a significant impact on the long-term performance; (2) deeper planning horizons can both diminish and amplify the impact; and (3) automatic methods can compute prior beliefs with consistent performance effects.

It is evident from this research that prior beliefs are a practically important and useful component of algorithms such as HBA. We hope that this work spurs further research on prior beliefs in the context of such algorithms. Specifically, this empirical study should be complemented with a theoretical analysis to solidify the observations made in this work.


For example, it is unclear if prior beliefs can be computed efficiently with useful performance guarantees. The LP-priors are a first attempt in this direction, since solving a LP-prior also provides a bound on the expected loss of whichever intrinsic metric is being optimised. However, as discussed previously, the LP formulation is a simplification of the true reasoning of HBA and so the bound may be incorrect.

	\section*{Acknowledgements}

This research was carried out while S.A. was a visiting student at Masdar Institute. S.A. wishes to thank Edmond~Awad and Nabil~Kenan. The authors acknowledge the support of the German National Academic Foundation, the Masdar Institute-MIT collaborative agreement under Flagship Project 13CAMA1, and the European Commission through SmartSociety Grant agreement no. 600854, under the programme FOCAS ICT-2011.9.10.

	\bibliographystyle{aaai}
	\bibliography{aaai15_prior}

\begin{thebibliography}{}

\bibitem[\protect\citeauthoryear{Albrecht and Ramamoorthy}{2012}]{ar2012}
Albrecht, S., and Ramamoorthy, S.
\newblock 2012.
\newblock Comparative evaluation of {MAL} algorithms in a diverse set of ad hoc
  team problems.
\newblock In {\em Proceedings of the 11th International Conference on
  Autonomous Agents and Multiagent Systems}, volume~1,  349--356.

\bibitem[\protect\citeauthoryear{Albrecht and Ramamoorthy}{2013}]{ar2013}
Albrecht, S., and Ramamoorthy, S.
\newblock 2013.
\newblock A game-theoretic model and best-response learning method for ad hoc
  coordination in multiagent systems (extended abstract).
\newblock In {\em Proceedings of the 12th International Conference on
  Autonomous Agents and Multiagent Systems},  1155--1156.

\bibitem[\protect\citeauthoryear{Albrecht and Ramamoorthy}{2014}]{ar2014}
Albrecht, S., and Ramamoorthy, S.
\newblock 2014.
\newblock On convergence and optimality of best-response learning with policy
  types in multiagent systems.
\newblock In {\em Proceedings of the 30th Conference on Uncertainty in
  Artificial Intelligence},  12--21.

\bibitem[\protect\citeauthoryear{Albrecht, Crandall, and
  Ramamoorthy}{2015}]{acr2015priorapp}
Albrecht, S.; Crandall, J.; and Ramamoorthy, S.
\newblock 2015.
\newblock An empirical study on the practical impact of prior beliefs over
  policy types -- {A}ppendix.
\newblock
  \newline\textit{http://rad.inf.ed.ac.uk/data/publications/2015/aaai15app.pdf}.

\bibitem[\protect\citeauthoryear{Barrett, Stone, and Kraus}{2011}]{bsk2011}
Barrett, S.; Stone, P.; and Kraus, S.
\newblock 2011.
\newblock Empirical evaluation of ad hoc teamwork in the pursuit domain.
\newblock In {\em Proceedings of the 10th International Conference on
  Autonomous Agents and Multiagent Systems}, volume~2,  567--574.

\bibitem[\protect\citeauthoryear{Bernardo}{1979}]{b1979}
Bernardo, J.
\newblock 1979.
\newblock Reference posterior distributions for {B}ayesian inference.
\newblock {\em Journal of the Royal Statistical Society. Series B
  (Methodological)} 41(2):113--147.

\bibitem[\protect\citeauthoryear{Bowling and McCracken}{2005}]{bm2005}
Bowling, M., and McCracken, P.
\newblock 2005.
\newblock Coordination and adaptation in impromptu teams.
\newblock In {\em Proceedings of the 20th National Conference on Artificial
  Intelligence}, volume~1,  53--58.

\bibitem[\protect\citeauthoryear{Brafman and Tennenholtz}{2003}]{bt2003}
Brafman, R., and Tennenholtz, M.
\newblock 2003.
\newblock R-max -- {A} general polynomial time algorithm for near-optimal
  reinforcement learning.
\newblock {\em Journal of Machine Learning Research} 3:213--231.

\bibitem[\protect\citeauthoryear{Brown}{1951}]{b1951}
Brown, G.
\newblock 1951.
\newblock Iterative solution of games by fictitious play.
\newblock {\em Activity analysis of production and allocation} 13(1):374--376.

\bibitem[\protect\citeauthoryear{Carberry}{2001}]{c2001}
Carberry, S.
\newblock 2001.
\newblock Techniques for plan recognition.
\newblock {\em User Modeling and User-Adapted Interaction} 11(1-2):31--48.

\bibitem[\protect\citeauthoryear{Carmel and Markovitch}{1999}]{cm1999}
Carmel, D., and Markovitch, S.
\newblock 1999.
\newblock Exploration strategies for model-based learning in multi-agent
  systems: Exploration strategies.
\newblock {\em Autonomous Agents and Multi-Agent Systems} 2(2):141--172.

\bibitem[\protect\citeauthoryear{Charniak and Goldman}{1993}]{cg1993}
Charniak, E., and Goldman, R.
\newblock 1993.
\newblock A {B}ayesian model of plan recognition.
\newblock {\em Artificial Intelligence} 64(1):53--79.

\bibitem[\protect\citeauthoryear{Claus and Boutilier}{1998}]{cb1998}
Claus, C., and Boutilier, C.
\newblock 1998.
\newblock The dynamics of reinforcement learning in cooperative multiagent
  systems.
\newblock In {\em Proceedings of the 15th National Conference on Artificial
  Intelligence},  746--752.

\bibitem[\protect\citeauthoryear{Cox, Shachat, and Walker}{2001}]{cjm2001}
Cox, J.; Shachat, J.; and Walker, M.
\newblock 2001.
\newblock An experiment to evaluate {B}ayesian learning of {N}ash equilibrium
  play.
\newblock {\em Games and Economic Behavior} 34(1):11--33.

\bibitem[\protect\citeauthoryear{Crandall}{2014}]{c2014}
Crandall, J.
\newblock 2014.
\newblock Towards minimizing disappointment in repeated games.
\newblock {\em Journal of Artificial Intelligence Research} 49:111--142.

\bibitem[\protect\citeauthoryear{De~Finetti}{2008}]{f2008}
De~Finetti, B.
\newblock 2008.
\newblock {\em Philosophical Lectures on Probability: collected, edited, and
  annotated by Alberto Mura}.
\newblock Springer.

\bibitem[\protect\citeauthoryear{Dekel, Fudenberg, and Levine}{2004}]{dfl2004}
Dekel, E.; Fudenberg, D.; and Levine, D.
\newblock 2004.
\newblock Learning to play {B}ayesian games.
\newblock {\em Games and Economic Behavior} 46(2):282--303.

\bibitem[\protect\citeauthoryear{Gmytrasiewicz and Doshi}{2005}]{gd2005}
Gmytrasiewicz, P., and Doshi, P.
\newblock 2005.
\newblock A framework for sequential planning in multiagent settings.
\newblock {\em Journal of Artificial Intelligence Research} 24(1):49--79.

\bibitem[\protect\citeauthoryear{Harsanyi}{1967}]{h1967}
Harsanyi, J.
\newblock 1967.
\newblock Games with incomplete information played by ``{B}ayesian'' players.
  {Part I. T}he basic model.
\newblock {\em Management Science} 14(3):159--182.

\bibitem[\protect\citeauthoryear{Holland}{1975}]{h1975}
Holland, J.
\newblock 1975.
\newblock {\em Adaptation in natural and artificial systems: An introductory
  analysis with applications to biology, control, and artificial intelligence}.
\newblock The MIT Press.

\bibitem[\protect\citeauthoryear{Jaynes}{1968}]{j1968}
Jaynes, E.
\newblock 1968.
\newblock Prior probabilities.
\newblock {\em IEEE Transactions on Systems Science and Cybernetics}
  4(3):227--241.

\bibitem[\protect\citeauthoryear{Jordan}{1991}]{j1991}
Jordan, J.
\newblock 1991.
\newblock Bayesian learning in normal form games.
\newblock {\em Games and Economic Behavior} 3(1):60--81.

\bibitem[\protect\citeauthoryear{Kalai and Lehrer}{1993}]{kl1993}
Kalai, E., and Lehrer, E.
\newblock 1993.
\newblock Rational learning leads to {N}ash equilibrium.
\newblock {\em Econometrica} 61(5):1019--1045.

\bibitem[\protect\citeauthoryear{Koza}{1992}]{k1992}
Koza, J.
\newblock 1992.
\newblock {\em Genetic programming: On the programming of computers by means of
  natural selection}.
\newblock The MIT Press.

\bibitem[\protect\citeauthoryear{Rapoport and Guyer}{1966}]{rg1966}
Rapoport, A., and Guyer, M.
\newblock 1966.
\newblock A taxonomy of $2 \times 2$ games.
\newblock {\em General Systems: Yearbook of the Society for General Systems
  Research} 11:203--214.

\end{thebibliography}

\end{document}